\documentclass[sigconf]{acmart}
\AtBeginDocument{%
  \providecommand\BibTeX{{%
    \normalfont B\kern-0.5em{\scshape i\kern-0.25em b}\kern-0.8em\TeX}}}

\copyrightyear{2022}
\acmYear{2022}
\setcopyright{acmlicensed}
\acmConference[MM '22]{Proceedings of the 30th ACM International Conference on Multimedia}{October 10--14, 2022}{Lisboa,Portugal}
\acmBooktitle{Proceedings of the 30th ACM International Conference on
Multimedia (MM '22), October 10--14, 2022, Lisboa, Portugal}
\acmPrice{15.00}
\acmDOI{10.1145/3503161.3548191}
\acmISBN{978-1-4503-9203-7/22/10}

\usepackage{multirow}
\usepackage{float}
\usepackage{bbding}
\usepackage{graphicx}
\usepackage{algorithm}
\usepackage{algorithmic}
\usepackage{booktabs}
\usepackage{subfigure}
\usepackage{enumitem}
\usepackage{lineno}
\usepackage{xspace}

\newcommand*{\eg}{\emph{e.g.}\@\xspace}
\newcommand*{\ie}{\emph{i.e.}\@\xspace}
\newcommand*{\etal}{\emph{et al.}\@\xspace}
\newcommand*{\etc}{\emph{etc.}\@\xspace}

\acmSubmissionID{mmfp1843}  
\begin{document}

\title{SlimSeg: Slimmable Semantic Segmentation with Boundary Supervision}


\author{Danna Xue}
\email{danna_xue@mail.nwpu.edu.cn}
\affiliation{
  \institution{Northwestern Polytechnical University \& Computer Vision Center, Universitat Autònoma de Barcelona}
  \city{Xi'an}
  \country{China}
  \postcode{710129}
}

\author{Fei Yang}
\email{fyang@cvc.uab.es}
\affiliation{
  \institution{Computer Vision Center, Universitat Autònoma de Barcelona}
%   \streetaddress{Edifici O, Campus UAB, 08193 Bellaterra}
  \city{Barcelona}
  \country{Spain}}

\author{Pei Wang}
\email{wangpei23@mail.nwpu.edu.cn}
\affiliation{%
  \institution{Northwestern Polytechnical University}
  \streetaddress{DongXiang Street, 1}
  \city{Xi'an}
  \country{China} 
}

\author{Luis Herranz}
\email{lherranz@cvc.uab.es}
\affiliation{
  \institution{Computer Vision Center, Universitat Autònoma de Barcelona}
  \streetaddress{Edifici O, Campus UAB, 08193 Bellaterra}
  \city{Barcelona}
  \country{Spain}}

\author{Jinqiu Sun}
\authornote{Corresponding author.}
\email{sunjinqiu@nwpu.edu.cn}
\affiliation{
  \institution{Northwestern Polytechnical University}
%   \streetaddress{DongXiang Street, 1}
  \city{Xi'an}
  \country{China} 
}

\author{Yu Zhu}
\email{yuzhu@nwpu.edu.cn}
\affiliation{
  \institution{Northwestern Polytechnical University}
%   \streetaddress{DongXiang Street, 1}
  \city{Xi'an}
  \country{China} 
}

\author{Yanning Zhang}
\email{ynzhang@nwpu.edu.cn}
\affiliation{
  \institution{Northwestern Polytechnical University}
%   \streetaddress{DongXiang Street, 1}
  \city{Xi'an}
  \country{China} 
}


\renewcommand{\shortauthors}{Danna Xue et al.}
\begin{abstract}
Accurate semantic segmentation models typically require significant computational resources, inhibiting their use in practical applications. Recent works rely on well-crafted lightweight models to achieve fast inference. However, these models cannot flexibly adapt to varying accuracy and efficiency requirements. In this paper, we propose a simple but effective slimmable semantic segmentation (SlimSeg) method, which can be executed at different capacities during inference depending on the desired accuracy-efficiency tradeoff. More specifically, we employ parametrized channel slimming by stepwise downward knowledge distillation during training. Motivated by the observation that the differences between segmentation results of each submodel are mainly near the semantic borders, we introduce an additional boundary guided semantic segmentation loss to further improve the performance of each submodel. We show that our proposed SlimSeg with various mainstream networks can produce flexible models that provide dynamic adjustment of computational cost and better performance than independent models. Extensive experiments on semantic segmentation benchmarks, Cityscapes and CamVid, demonstrate the generalization ability of our framework.

\end{abstract}





\maketitle

%===============================================================

\section{Introduction}

Semantic segmentation predicts the semantic category corresponding to each pixel in an image. Various applications have benefited from advances towards more accurate results, such as autonomous driving \cite{li2020dabnet, li2019dfanet, yu2021bisenet, hong2021deep, gao2021rethink, zhang2019customizable, li2020humans, lin2020graph, chen2020fasterseg, zhang2021dcnas, yang2021real, li2020semantic, SiZL20}, image synthesis and manipulation \cite{wang2018high, park2019gaugan}, and medical imaging \cite{qin2021efficient, li2021semantic}. Based on the pioneering fully convolutional network \cite{long2015fully}, previous studies have made important achievements by greatly increasing the performance on various challenging semantic segmentation benchmarks \cite{cordts2016cityscapes,brostow2008segmentation,zhou2017scene, Everingham10}. Despite their superiority, these powerful models, built upon heavy deep neural networks, suffer from the low inference speed and strict requirements for computing devices.

% \vspace{4mm}
\begin{table}[h]
\caption{The FLOPs and number of parameters of semantic segmentation networks (except for the backbone) and their proportions of the whole model, with image size 1024$\times$2048. }
% \vspace{-2mm}
\label{table:param}
\setlength{\tabcolsep}{0.9mm}{
\begin{tabular}{c|c|c|c|c}
\toprule
Networks & \multicolumn{2}{c|}{SFNet \cite{li2020semantic}}   & \multicolumn{2}{c}{DeepLabv3+ \cite{chen2018encoder}} \\  \midrule
Backbone & ResNet50     & ResNet18     & ResNet50       & MobileNetv2   \\  \midrule
GFLOPs                  & 436.3 | 72\% & 107.5 | 55\% & 663.5 | 45\%   & 6.3 | 34\%   \\
Params                    & 7.7M | 25\%  & 1.5M | 12\%  & 16.8M | 40\%   & 2.7M | 59.3\%   \\ \bottomrule
% \vspace{-3mm}
\end{tabular}}
\end{table}

Most of the existing works mainly address efficient semantic segmentation through (i) designing compact backbone architectures \cite{treml2016speeding, romera2017erfnet, zhao2018icnet,  li2020dabnet, li2019dfanet, yu2021bisenet, hong2021deep, gao2021rethink}, (ii) effective model compression methods \cite{paszke2016enet, li2019partial, zhang2019customizable, li2020humans, lin2020graph, chen2020fasterseg}, (iii) exploiting reliable context and boundary information \cite{li2020semantic,liu2020structured, wang2020ifvd, SiZL20, yang2021real}. 
However, those methods mainly speed up the inference with fixed network structures, while in practice, the equipped resources are quite different across diverse devices. Even for the same device, the availability of hardware resources varies over time. Suppose we want to switch between models of different sizes according to the ideal accuracy-efficiency tradeoff. One straightforward way is to train multiple independent models with different structures and parameters and load a specific one during inference. However, it requires a longer training time and more memory for storage. Unlike previous works, we focus on improving the flexibility of the semantic segmentation model. 
% Our model can adapt to available resources by adjusting its capacity within the same shared model parameters.

The recent work \cite{yu2018slimmable} proposed a slimmable neural network that can adjust the width of the network for different inference speeds. However, they mainly focus on image classification and only apply their slimmable models as backbones on instance segmentation tasks, while the other parts (\eg, the decoder) are non-slimmable. Due to the resolution of the output image, even if a relatively simple structure is used in the decoder part, including up-sampling and multi-level feature aggregation \etc, the decoder still requires a large amount of computation during inference. We show the computation cost (in FLOPs) and the number of parameters of several mainstream segmentation models, including SFNet \cite{li2020semantic} and DeepLabv3+ \cite{chen2018encoder}, in Table \ref{table:param}. In these models, the Pyramid Pooling Module (PPM) \cite{zhao2017pyramid} and the decoder account for more than one-third of the overall calculation, while the parameters for most of them are the minority of the whole model. Based on \cite{yu2018slimmable}, we focus on semantic segmentation and aim to lower computational cost from the perspective of reducing the overall size of the network, rather than just backbones. Motivated by this, we propose a slimmable semantic segmentation network (SlimSeg) that leverages the slimming mechanism to dynamically adjust the channel of features on every single layer. The network's capacity can be switched with the size of width according to the computational requirements, thereby controlling the trade-off between accuracy and inference time. In addition, we apply stepwise downward inplace distillation for training smaller subnetworks, which means that smaller subnetworks are learned from the larger ones. This leads to consistent results between different submodels.

Moreover, we also found that the differences between the predicted results of slimmable subnetworks with different widths mainly exist along the semantic boundaries. Previous works \cite{zhu2019improving, yuan2020segfix} also report that most existing segmentation models fail to make right predictions along the semantic boundaries. To further improve the segmentation quality on the boundary and narrow the accuracy gap between each subnetwork, we introduce a semantic boundary detection head on the low-level features and additional supervision named semantic boundary guided loss. This loss leverages the predicted boundaries as guidance to calculate a weighted bootstrapped cross-entropy. The boundary detection head can be removed during inference, so it does not introduce any additional computation. 

Our SlimSeg is a general scheme that can adapt the existing segmentation models to width switchable models without any new structural design. The experimental results on Cityscapes \cite{cordts2016cityscapes} and CamVid \cite{brostow2008segmentation} based on SFNet \cite{li2020semantic} and DeepLabv3+ \cite{chen2018encoder} demonstrate the slimmable model has comparable accuracy to independent models. Furthermore, our method shows higher accuracy on smaller subnetworks with the stepwise downward distillation and proposed boundary guided loss. The contributions are summarized as follows:

\begin{itemize}
% \vspace{-4mm}
 \item We propose a simple but effective slimmable semantic segmentation method (SlimSeg) which can adjust the capacity of the model depending on the desired trade-off between accuracy and efficiency.
 \item We present the boundary supervision, including a low-level boundary detection head and a boundary guided loss to improve the accuracy of semantic segmentation in boundary regions, especially for the smaller subnetworks.
 \item Extensive experiments and analysis indicate the efficacy and generalization ability of our proposed method, both quantitatively and qualitatively. 
%Our method achieves comparable results on both accuracy and speed on semanitc segmentation benchmarks, Cityscapes and CamVid.

\end{itemize}

%===============================================================
% \vspace{-4mm}

\section{Related Works}

\subsection{Generic Semantic Segmentation}

A typical semantic segmentation architecture generally includes two parts: encoder and decoder. The encoder module extracts image features through convolution and downsampling. Generally, the encoder is adapted from image classification models trained on ImageNet \cite{deng2009imagenet}, such as VGG19 \cite{simonyan2014very}, ResNet \cite{he2016deep}, \etc Since semantic segmentation conduct pixel-level classification, the typical fully connected layers are replaced by convolutional layers \cite{long2015fully}. To utilize the global context, the Pyramid Pooling Module (PPM) \cite{zhao2017pyramid, chen2018encoder} is employed to increase the receptive field without an increase in parameters. However, massive computations are introduced by PPM and other feature fusion modules performed on high-resolution features neighbor to the output. To pursue better global and local feature fusion, models \cite{sun2019high, zheng2021rethinking} based on more powerful backbones, such as HRNet \cite{wang2020deep} and ViT \cite{dosovitskiy2020image}, have been proposed. These models have achieved higher accuracy, but are limited by the hardware requirements in practice. Our approach takes advantage of the sophisticated models and achieves variable capacity through width slimming, enabling fast inference while maintaining accuracy.

% \vspace{-4mm}
\subsection{Efficient Semantic Segmentation}
Efficient semantic segmentation needs to consider both accuracy and computational cost. Existing methods trade accuracy and speed along three different lines.

\textbf{Hand-crafted compact backbone architecture.}
An effective backbone can greatly improve the upper bound of performance. The works \cite{treml2016speeding, romera2017erfnet, zhao2018icnet, li2020dabnet, li2019dfanet, yu2021bisenet, hong2021deep, gao2021rethink} design lightweight backbone architectures from scratch to pursue more efficient inference. Some works \cite{zhao2018icnet, li2019dfanet, hong2021deep} devised multiscale image cascades and feature fusion mechanisms to achieve a good accuracy-speed trade-off. Others \cite{li2020dabnet} improve existing network layers to create sufficient receptive field and densely utilize the contextual information. BiSeNet \cite{yu2021bisenet} introduced a shallow spatial branch to process full resolution images while learning context information by a deep branch.

\textbf{Machine-driven architecture optimization.}
Neural Architecture Search (NAS) \cite{zoph2016neural} is an effective technique to switch the labor-intensive architecture design to an automatic machine-driven optimization process, and this technique has been applied to semantic segmentation in recent years. From repeated cell structure \cite{paszke2016enet, zhang2019customizable} to more flexible network structure \cite{li2020humans}, different types of network (\eg, graph convolution network \cite{lin2020graph}), or explicitly taking latency into consideration \cite{li2019partial, chen2020fasterseg}. FasterSeg \cite{chen2020fasterseg} introduces the teacher-student co-searching and flexible multi-resolution branches aggregation structure. Although the latitude of the search space is continuously improved \cite{zhang2021dcnas}, it still requires longer training time and more effective search strategies.

\textbf{Feature mining and aggregation.}
By exploiting the potential of existing lightweight models, rather than building new architectures, these methods learn more favorable context information. Knowledge distillation \cite{hinton2015distilling} has shown its effectiveness on segmentation tasks by improving the accuracy of a lightweight student model and speed-up its convergence by transferring learned knowledge from a sophisticated teacher network. Liu \etal \cite{liu2020structured, wang2020ifvd} provide a comprehensive analysis of feature distillation at different levels, from various cumbersome models to compact models. Others investigate multi-level feature aggregation to alleviate the side effects of up and down sampling \cite{li2020semantic} or enlarge the receptive field of lightweight networks \cite{SiZL20, yang2021real}.

Although these efficient semantic segmentation approaches improve the accuracy-efficiency tradeoff from different perspectives, the resulting model is still limited to fixed size and operating at a single tradeoff. Unlike these methods, we enable adjustable computation with one single model and ensures good accuracy for each submodel of different size.

% \vspace{-2mm}
\subsection{Dynamic Neural Networks}
Dynamic neural networks \cite{han2021dynamic} reduce average inference cost by adaptively changing characteristics of the computational graph, including the resolution, depth, and width. Reducing the \textbf{resolution} of the input image is the most straightforward way to lower computational costs. For images with relatively simple context, equivalent prediction accuracy can be achieved with lower resolutions. Some works \cite{wang2020resolution, yang2020resolution, zhu2021dynamic} propose parallel training for multi-resolution inference with a single model. Networks with dynamic \textbf{depth} speed up inference by skipping residual blocks adaptively \cite{wang2018skipnet, veit2018convolutional, li2020learning} or early exiting when shallower subnetworks have high enough confidence \cite{GaoHuang2018MultiScaleDN, yang2020resolution, kouris2021multi}. The number of feature channels, \ie \textbf{width}, is also a key factor of efficiency. One way of enabling various channel inference is dynamic pruning. By identifying and skipping the insignificant channels during inference \cite{hua2019channel, gao2018dynamic, li2021dynamic} or training a hypernetwork to select the filters \cite{chen2019you}, the channel complexity can be lessened. Moreover, \cite{yu2018slimmable, yu2019universally, yu2019autoslim} propose slimmable neural networks with embedded submodels sharing parameters that are executable at different widths, allowing immediate and adaptive accuracy-efficiency trade-offs at runtime. Based on the success of slimmable neural network, Liang \etal \cite{li2021dynamic} improve the hardware efficiency by introducing a dynamic slimming gate that adaptively adjusts the network width with negligible extra computation cost. 
Although dynamic neural networks have shown their effectiveness on strategically allocating appropriate computational resources, most works still focus on image classification and some other low-level vision tasks, such as image compression \cite{yang2021slimmable}, denoising \cite{jiang2021dynamic} and image generation \cite{hou2020slimmable}. Different from previous works, we study dynamic semantic segmentation models through our analysis.

%===============================================================

\section{Method}
% In this section, we first introduce the pipeline of our proposed framework in Sec. 3.1. Then we describe the stepwise downward knowledge distillation and the semantic boundary guided loss in the Sec. 3.2 and 3.3, respectively. 

\subsection{Slimmable Segmentation Framework}
Image semantic segmentation requires assigning a category label to each pixel in the image from several semantic categories. Given an image $x$, a segmentation network $\mathcal{S}$ parameterized by $\theta$ implements a mapping $p=\mathcal{S}(x;\theta)$, where each spatial element of $p$ is a probability vector indicating the probability of each semantic category, from which the most probable is selected. Ideally, it should correspond to the category indicated in the corresponding ground truth segmentation map $y$ (coded as one-hot probability vectors per pixel). During training, the loss minimized is the cross-entropy  $\mathcal{L}_{CE}\left(p,y;\theta\right)$ between the predicted probability and the ideal one-hot label. In practice, this loss is averaged over the pixels in the image and the image-segmentation pairs $\left(x,y\right)$ in the training dataset.

\begin{figure*}[h]
  \includegraphics[scale=0.64]{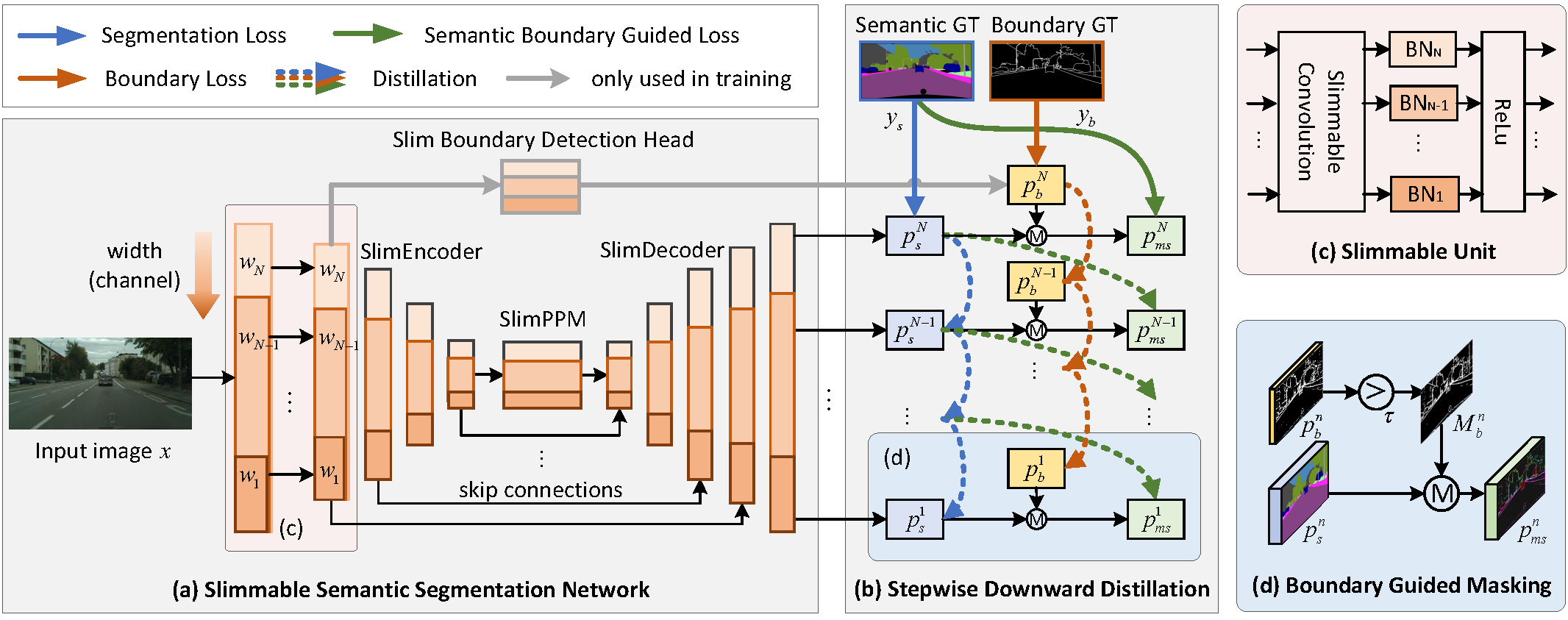}
%   \vspace{-1.5mm}
  \caption{Overview of our slimmable semantic segmentation framework. (a) The whole network, including the encoder, PPM, decoder and boundary detection head, is slimmable. The boundary detection head can be removed during inference. (c) Each slimmable unit includes a slimmable convolution layer, independent BNs for each width and a ReLu layer. (b) The largest network with width $w_N$ is supervised by the ground truth labels, and the smaller models with width $w_n$ are learned from larger models with width $w_{n+1}$ by stepwise distillation. (d) The predicted boundaries are used to generate boundary masked probability maps for calculating the boundary guided loss.}
%   \vspace{-1.5mm}
  \label{fig:framework}
\end{figure*}

In this work, we propose a flexible semantic segmentation framework, named as SlimSeg, which can adapt its model capacity during inference via the slimming mechanism to accommodate various levels of computing power. More specifically, we define different sets of widths (\ie number of channels in each convolutional layer) of the segmentation network. Thus, the segmentation network contains $N$ subnetworks with parameters $\left \{ \theta_{w_{1}}, \theta_{w_{2}}, ... ,\theta_{w_{N}} \right \}$ with $N$ increasing widths $w_{1}<w_{2}<... <w_{N}$, respectively. For every convolutional layer implementing slimming, the parameters are built as subsets of larger (sub)networks as $\theta_{w_{1}} \subset \theta_{w_{2}} \subset... \subset \theta_{w_{N}}= \theta$. Then, the objective of our task becomes optimizing all the subnetworks with $\sum ^{N}_{n=1} \mathcal{L}_{CE}\left(p^n,y;\theta_{w_n}\right)$, where $p^n$ is the predicted category probability vector of the $n^{th}$ subnetworks with parameters $\theta_{w_n}$. The loss is also averaged over pixels and training data, and then minimized over the parameters $\theta$. Note that we could also replace the (one-hot) ground truth label $y$ with the soft label $p_{n'}$ predicted by larger subnetworks to distill its knowledge. We describe our loss functions in more detail in Section 3.2 and 3.3. Henceforth, we will also omit the explicit dependencies on the model parameters for the sake of simplicity.

The overall pipeline of our SlimSeg is illustrated in Figure \ref{fig:framework}. We deploy width slimming on the entire network, including the encoder for feature extraction, the Pyramid Pooling Module \cite{zhao2017pyramid} and the decoder for feature aggregation and classification. The number of channels is adjusted through the slimmable convolutional layer \cite{yu2018slimmable}, which produces different output feature channels by adjusting the number of filters. The slimmable convolution will result in a different output feature distribution. Following \cite{yu2018slimmable}, we use independent batch normalization (BN) layers for each width, which only introduces very few parameters to the overall model.

\subsection{Stepwise Downward Distillation}

To utilize the knowledge learned by large submodels to guide the learning of the smaller submodels, we apply inplace knowledge distillation from larger (sub)networks to smaller ones. Unlike previous knowledge distillation on segmentation \cite{liu2020structured, wang2020ifvd}, we do not learn from an already trained (fixed) sophisticated model to improve another independent compact model. We introduce stepwise downward inplace distillation, where class probabilities estimated from the larger subnetwork are used as soft targets for training the next smaller subnetwork. The largest subnetwork is supervised by the ground truth labels. Note that the parameters of a smaller subnetwork are also a subset of larger ones, which means that the smallest subnetwork will learn the most important features implicitly to guarantee the accuracy of larger submodels. This leads to the following loss function:

% \vspace{-2.5mm}
\begin{equation}
\mathcal{L}_{seg} = \mathcal{L}_{CE}(p^{N}_{s},y_{s})+\sum_{n=1}^{N-1}\mathcal{L}_{KD}(p^{n}_{s},p^{n+1}_{s}),
\label{loss_seg}
\end{equation}

\noindent where $\mathcal{L}_{CE}$ denotes the cross entropy loss, and $p^{n}_{s}$, $y_{s}$ are the segmentation probability map predicted by the $n^{th}$ submodel and the ground truth semantic label, respectively. 
Instead of computing the Kullback-Leibler divergence between two probabilities, we use soft target cross-entropy loss (we denote it as $\mathcal{L}_{KD}$ to distinguish it from $\mathcal{L}_{CE}$, which applied with ground truth supervision). We found that the cross-entropy between two probabilities is more stable during training than the Kullback-Leibler divergence, which is also a common setting for the knowledge distillation in \cite{yu2018slimmable, yu2019universally, yu2019autoslim, liu2020structured}. 

In practice, stopping the gradients of the supervising tensor predicted by the larger width is necessary, so that the loss of a subnetwork will never back-propagate through the computation graph to larger subnetworks. We performed experiments on the effectiveness of distillation and the type of optimal teachers. The results show that using the probability map predicted by previous subnetworks as the soft target can lead to better performance. For more details, see Section \ref{ablation}.

% \vspace{-2mm}
\subsection{Semantic Boundary Guided Loss}

\begin{figure}[h]
  \centering
  \includegraphics[scale=0.68]{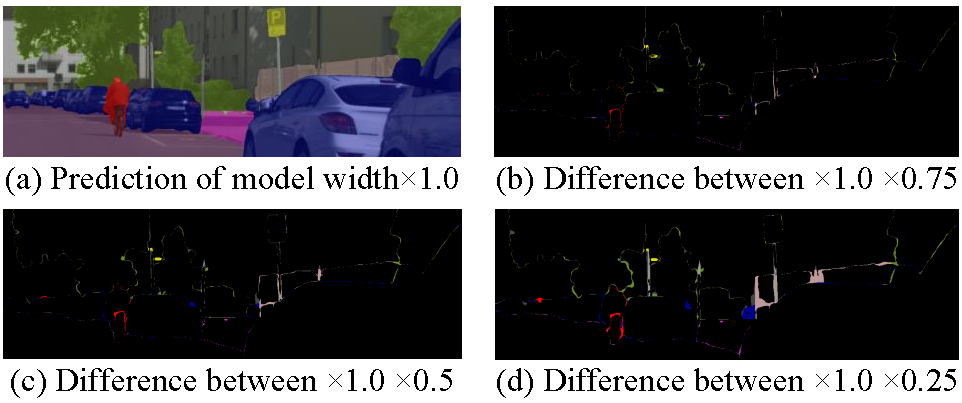}
%   \vspace{-2mm}
  \caption{Difference between submodels. (a) Predicted semantic map of the $\times1.0$ model. (b)-(d) Difference map between the smaller submodels and the $\times1.0$ model, where the consistent (inconsistent) predicted pixels are shown as black background (ground truth color codes). Better view in color.}
  \label{fig:edge_analysis}
% \vspace{-2mm}
\end{figure}

Based on the training framework and distillation method presented above, we can already obtain varying amounts of computation of multiple subnetworks with partially shared parameters. To further improve the performance, especially for the smaller subnetworks, we compare the semantic labels predicted by different subnetworks trained only with the loss $\mathcal{L}_{seg}$. As illustrated in Figure \ref{fig:edge_analysis}, the differences between the segmentation results of subnetworks with different widths are mainly near the borders between different semantic categories. Moreover, as the width decreases, the gap between the predictions gets larger.

\vspace{-2mm}

Motivated by this observation, we introduce extra boundary supervision to improve the accuracy in those regions, especially for small subnetworks. Specifically, we introduce an additional boundary detection head with a simple structure, including a slimmable unit (Figure \ref{fig:framework} (c)) and a slimmable convolution layer with kernel size 1 followed by a sigmoid layer, on the low-level features. The output of this head $p^{N}_{b}$ is supervised by the binary boundary masks generated by the semantic segmentation ground truth labels $y_{b}$. The pixels within 3 pixels from the semantic border are marked as boundary regions. We apply binary cross-entropy loss to constrain boundary detection with:

% \vspace{-2mm}
\begin{equation}
\mathcal{L}_{b} = \mathcal{L}_{BCE}(p^{N}_{b}, y_{b})+\sum_{n=1}^{N-1}\mathcal{L}_{KD}(p^{n}_{b}, p^{n+1}_{b}),
\end{equation}

\noindent where we also leverage knowledge distillation to subnetworks with the soft boundary labels predicted by the larger one, except for the largest width that uses the boundary ground truth $y_{b}$. Unlike \cite{ding2019boundary}, our boundary detection head is used only on training and can be removed during inference, so it does not introduce any extra computation. The head helps enhance the low-level features of boundary regions.

Besides, the estimated boundary also perform as a reference to resample the misclassified pixels on the border to calculate the boundary guided segmentation loss, which can be regarded as a hard sample mining strategy. As shown in Figure \ref{fig:framework} (d), taking the boundary probability map $p_{b}$ predicted by the boundary detection head, we generate a confidence binary mask $M_{b}$ to locate those pixels which might be situated near to semantic boundaries:
% \vspace{-0.5mm}
\begin{equation}
M_{b}(u,v)=\left\{\begin{matrix}
valid, & p_{b}(u,v)>\tau \\ 
invalid, & otherwise
\end{matrix}\right ..
\end{equation}

\noindent The values in $M_{b}$ are element-wise calculated by comparing the boundary confidential score $p_{b}$ at each location $(u,v)$ with a predefined threshold $\tau$. We empirically set $\tau$ to 0.7 in our experiments. Only valid pixels are included in the loss calculation. Similar to $\mathcal{L}_{seg}$, the cross-entropy loss and the knowledge distillation loss of the masked semantic probabilities $p^{n}_{ms} = M^{n}_{b}( p^{n}_{s}), n \in \left \{1,2,\cdots, N  \right \}$ are calculated with:

\begin{equation}
\mathcal{L}_{g} = \mathcal{L}_{CE}(p^{N}_{ms}, y_{s}) +\sum_{n=1}^{N-1}\mathcal{L}_{KD}(p^{n}_{ms}, p^{n+1}_{s}).
\end{equation}

Then, the loss function for training our SlimSeg is calculated as a summation of the semantic segmentation loss $\mathcal{L}_{seg}$, boundary detection loss $\mathcal{L}_{b}$ and the boundary guided segmentation loss $\mathcal{L}_{g}$:
\begin{equation}
\mathcal{L}_{full} = \mathcal{L}_{seg} + \lambda_{1}\mathcal{L}_{b} + \lambda_{2}\mathcal{L}_{g} \label{eq:fullloss}
\end{equation}
\noindent where $\lambda_{1}, \lambda_{2}$ are hyperparameters, which are set to 10 and 1 in our experiments, respectively.

Finally, to clarify the training procedure of our proposed SlimSeg, we provide a Pytorch-style pseudo-code in Algorithm \ref{algorithm}.

% \begin{minipage}{7cm}
\begin{algorithm}
    \caption{Slimmable semantic segmentation}
    \label{algorithm}
    \begin{algorithmic}[1]
        \ENSURE{Dataset $\mathcal{D}$, width list $ \mathcal{W} = \left \{ w_{1},w_{2},...,w_{N} \right \}$}
        \REQUIRE{Slimmable semantic segmentation network $\mathcal{S}$}
        \FOR{$i=1,2,...,iteration$}
        \STATE{Get a mini-batch of image $x$, semantic label $y_{s}$, boundary label $y_{b}$ from $\mathcal{D}$.}
        \STATE{Clear gradients of weights, $optimizer.zerograd()$.} 
            \FOR{$w$ in $sorted(\mathcal{W}, reverse=True)$}
            \STATE{Switch the BN layers to current width.}
            \STATE{Execute current subnetwork, ${p}_{s}, {p}_{b}=\mathcal{S}(x;\theta_{w})$.}
            \STATE{Compute the masked probability, ${p}_{ms}=M_{b} (p_{s})$.}
            \IF {$w=w_{N}$}
                \STATE{Compute loss with ground truth, \\ $loss=CE(p_{s}, y_{s})+ BCE(p_{b}, y_{b})+CE(p_{ms}, y_{s})$.}
            \ELSE
                % \STATE{Compute target from Teachers $\mathcal{T}$}
                \STATE{Compute distillation loss,\\
                $loss=KD(p_{s}, y^{t}_{s})+ KD(p_{b}, y^{t}_{b})+ KD(p_{ms}, y^{t}_{s})$.}
            \ENDIF
            \IF {$w>w_{1}$}
                \STATE{Save predicted probability $p_{s}$, $p_{b}$  as teachers $y^{t}_{s}$, $y^{t}_{b}$.}
            \ENDIF
            \STATE{Compute gradients, $loss.backward()$.}
            \ENDFOR
            \STATE{Update weights, $optimizer.step()$.}
        \ENDFOR
        \RETURN  $\mathcal{S}$
    \end{algorithmic}
\end{algorithm}
% \end{minipage}

%===============================================================
\section{Experiments}

% In this section, we first describe the training datasets and implementation details. Then, we demonstrate the effectiveness of the design of SlimSeg through extensive ablation experiments. Finally, we compare our slimmable segmentation network with other state-of-the-art real-time semantic segmentation methods and discuss the impact of our proposed approach.

\subsection{Benchmarks and Evaluation Metrics}
% We implement our method on two standard semantic segmentation benchmarks: Cityscapes \cite{cordts2016cityscapes} and CamVid \cite{brostow2008segmentation} to evaluate the effectiveness of the proposed method. 
\textbf{Cityscapes. } Cityscapes \cite{cordts2016cityscapes} is a first-person perspective street-scene dataset with 19 semantic categories, 5000 fine annotated images with 2,975, 500 and 1,525 images for training, validation and testing, respectively. The high resolution of the images (1024$\times$2048 pixels) poses a great challenge to real-time semantic segmentation. For a fair comparison, we only use the fine annotated images for training.

\noindent \textbf{CamVid. } CamVid \cite{brostow2008segmentation} is a road scene dataset from the perspective of a driving automobile. It consists of 367, 101 and 233 images for training, validation and testing with resolution 720$\times$960. Following the pioneering work \cite{yu2021bisenet,fan2021rethinking}, we use the subset of 11 semantic classes from the 32 provided categories for a fair comparison with existing methods. The pixels out of the selected classes are ignored.

\noindent \textbf{Evaluation Metrics. } For quantitative evaluation, we report the mean of class-wise intersection-over-union (mIoU) for accuracy comparison. The floating-point operations per second (FLOPs) and frames per second (FPS) are adopted for efficiency comparison. Besides, we also give the number of parameters for model size. 
% Note that the units of measurement for FLOPs and number of parameters are GMac and M, respectively.

\subsection{Implementation Details}

\noindent \textbf{Training. } We use the stochastic gradient descent (SGD) algorithm to train our models with the batch size of 8, stochastic momentum of 0.9 and weight decay of 5e-4. As a common practice, the “poly” learning rate strategy in which the initial rate is multiplied by $\left ( 1-\frac{iter}{iter_{max}} \right )^{power}$ at each iteration with the power of 0.9. All the models are trained for 100K iterations with an initial learning rate of 0.01 and Online Hard Example Mining (OHEM) \cite{lin2017feature} on two NVIDIA GeForce 3090Ti GPUs   Data augmentation includes random horizontal flip, random resizing with the scale range of [0.5, 2.0], and random cropping to $768\times768$ for Cityscapes and $720\times720$ for CamVid. 

\noindent \textbf{Inference. } For inference, we use the whole image as an input to report performance, unless explicitly mentioned. Evaluation tricks such as sliding window inference and multiscale testing are not adopted. The measurement of inference time is executed on a single NVIDIA GeForce 2080Ti with CUDA 10.1, CUDNN 7.0, and we report the FPS without TensorRT acceleration.

\noindent \textbf{Architectures. } We conduct the experiments based on two mainstream semantic segmentation networks: SFNet \cite{li2020semantic} and DeepLabv3+ \cite{chen2018encoder}. SFNet is based on the Feature Pyramid Network \cite{lin2017feature} architecture with a backbone network pre-trained on ImageNet classification \cite{deng2009imagenet} as encoder, a pyramid pooling module and a decoder aggregating multi-level features from the encoder. Similarly, DeepLabv3+ \cite{chen2018encoder} includes a feature encoder, an atrous spatial pyramid pooling module and a simple decoder with only several convolutional layers and upsampling. For SFNet, we use the slimmable ResNet50 \cite{yu2018slimmable} pre-trained on ImageNet \cite{deng2009imagenet}, and slimmable ResNet18, DFNetV1, DFNetV2 \cite{li2019partial} without pre-training as encoder. For DeepLabv3+, we report the results using the slimmable ResNet50 and MobileNetv2 \cite{yu2018slimmable} (both are pre-trained on ImageNet) as encoder. The input of the boundary detection head is the low level features output by the second stage of the backbones. The resolution of the input features are down-sampled 4 times compared to the original image. We apply four width multipliers $[0.25, 0.5, 0.75, 1.0]$ in our experiments, except for Deeplabv3+-MobileNetv2 with $[0.35, 0.5, 0.75, 1.0]$.

\vspace{-3mm}
\subsection{Ablation Study}
\label{ablation}
We conduct ablation experiments to validate the effectiveness of our width slimming training scheme, knowledge distillation method and the proposed boundary guided loss.

\textbf{Width Slimming Training Scheme. } 
We compare the slimmable model with their independently trained counterparts to demonstrate the effectiveness of the width slimming segmentation training scheme. The independent models have the same architecture as the slimmable subnetworks, but can only operate on a single width. Note that both the independent and slimmable models are trained with the loss $\mathcal{L}_{full}$ in Eq.\ref{eq:fullloss} for fair comparison, and the independent models are supervised by ground truth. We report the mIoU, number of parameters (M) and FLOPs (GMac) in Table \ref{tab:main}. The slimmable models outperform the independent models of all width on SFNet (ResNet50, ResNet18) and DeepLabv3+ (ResNet50, MobileNetv2), while for SFNet (DFNetv, DFNetv2), the larger independent models are better than the slimmable one. We think this is because DFNet \cite{howard2019searching} is a compact backbone designed for best speed accuracy trade-off by neural architecture search, which have very little space to be compressed. Therefore, the gap between slimmable SFNet-DFNets submodels with different widths is also larger than ResNets. In terms of the amount of computation, with about 56\% of the whole FLOPs, the submodel with width$\times0.75$ achieves comparable performance as the full model. Besides, a slimmable model saves about 50\% memories for storing the parameters compared with several independent models, and number will increase if we have more switchable width.
% More details are shown in the supplementary materials.

\begin{table}[h]
\caption{Comparison of independent and slimmable models on Cityscapes \textit{val}. Bold numbers indicate the better mIoUs. }
\label{tab:main}
% \scalebox{1.0}{
\setlength{\tabcolsep}{0.9mm}{
\begin{tabular}{c|l|c|c|c|c|c}
\toprule
\multirow{2}{*}{Network} & \multirow{2}{*}{Width} & \multicolumn{2}{c|}{Independent} & \multicolumn{2}{c|}{Slimmable} & \multirow{2}{*}{FLOPs} \\ %\cline{3-6}
 &  & mIoU & Param & mIoU & Param &  \\ \midrule
\multirow{4}{*}{\begin{tabular}[c]{@{}c@{}}SFNet\\ ResNet50\end{tabular}} & $\times 1.0$ &78.3  & 31.20 &\textbf{78.4} (0.1$\uparrow$) &\multirow{4}{*}{31.29} & 607.9  \\
 & $\times 0.75$ &77.3 &17.57  &\textbf{77.9} (0.6$\uparrow$)   &   & 343.4  \\
 & $\times 0.5$ &76.3  &7.82  &\textbf{77.4} (1.1$\uparrow$)  &  & 153.9 \\
 & $\times 0.25$ &73.2  &1.97  &\textbf{74.4} (1.2$\uparrow$)  &  & 39.4 \\\midrule

\multirow{4}{*}{\begin{tabular}[c]{@{}c@{}}SFNet\\ ResNet18\end{tabular}} & $\times 1.0$ &75.0  &12.87    &\textbf{75.6} (0.6$\uparrow$) &\multirow{4}{*}{12.89} & 243.4 \\
 & $\times 0.75$ &74.0  &7.24  &\textbf{74.8} (0.8$\uparrow$)  &  &137.4  \\
 & $\times 0.5$ &71.4  &3.22  &\textbf{72.5} (1.1$\uparrow$)  &  & 61.5 \\
 & $\times 0.25$ &65.5  &0.79  &\textbf{67.3} (1.8$\uparrow$)  &  &15.7  \\\midrule
\multirow{4}{*}{\begin{tabular}[c]{@{}c@{}}SFNet\\ DFNetv2\end{tabular}} & $\times 1.0$ &\textbf{73.6}  &17.88  &73.1 (0.5$\downarrow$)  & \multirow{4}{*}{17.91} & 80.2 \\
 & $\times 0.75$ &\textbf{71.4} &10.06  &71.1 (0.3$\downarrow$)  &  & 45.2 \\
 & $\times 0.5$ &\textbf{70.0} &4.48  &69.8 (0.2$\downarrow$)  &  & 20.2 \\
 & $\times 0.25$ &62.5  &1.12  &\textbf{64.2} (1.7$\uparrow$)  &  & 5.2 \\\midrule
\multirow{4}{*}{\begin{tabular}[c]{@{}c@{}}SFNet\\ DFNetv1\end{tabular}} & $\times 1.0$ &\textbf{70.0}  &8.42  &69.4 (0.6$\downarrow$)  & \multirow{4}{*}{8.44} & 32.8 \\
 & $\times 0.75$ &\textbf{67.8}  &4.74   &67.0 (0.8$\downarrow$) &  & 18.6 \\
 & $\times 0.5$ &65.0  &2.11  &\textbf{65.3} (0.3$\uparrow$) &   & 8.4 \\
 & $\times 0.25$ &57.8  &0.52  &\textbf{59.8} (2.0$\uparrow$) &   & 2.2 \\ \midrule
\multirow{4}{*}{\begin{tabular}[c]{@{}c@{}}DeepLabv3+\\ ResNet50\end{tabular}} & $\times 1.0$ &78.0  &40.35  & \textbf{78.4} (0.4$\uparrow$)  & \multirow{4}{*}{40.44} & 1463 \\
& $\times 0.75$ &77.6  &22.71  & \textbf{78.2} (0.6$\uparrow$)   &   & 824.3 \\
 & $\times 0.5$ &76.7  &10.11  & \textbf{77.6} (0.9$\uparrow$)   &  & 347.6 \\
 & $\times 0.25$ &74.0  &2.54  & \textbf{75.6} (1.6$\uparrow$)   &   & 92.9 \\\midrule
 \multirow{4}{*}{\begin{tabular}[c]{@{}c@{}}DeepLabv3+\\ MobileNetv2\end{tabular}} & $\times 1.0$ &66.9  &4.53  &\textbf{67.9} (1.0$\uparrow$)  & \multirow{4}{*}{4.58} &18.5\\
 & $\times 0.75$ &63.3  &2.57  &\textbf{67.0} (3.7$\uparrow$)  &  &12.2  \\
 & $\times 0.5$ &58.6  &1.16  &\textbf{64.3} (5.7$\uparrow$)  &  & 5.7 \\
 & $\times 0.35$ &56.1  &0.57  &\textbf{61.1} (5.0$\uparrow$)  &  &3.3  \\ \bottomrule
\end{tabular}}
% }
\vspace{-2mm}
\end{table}

\textbf{Stepwise Downward Distillation. }
To make the most of the knowledge learned by larger submodels, we test different distillation settings and demonstrate the effectiveness of our distillation method.

\textit{Does inplace knowledge distillation work? }
We compare the mIoUs of training the slimmable model with and without stepwise downward distillation in Table \ref{tab:distillation}. For the smallest subnetwork with width$\times0.25$, the mIoUs consistently improve with distillation under all combinations of loss functions. With the distillation strategy proposed by our work, mIoUs improve on all subnetworks, and among them, the smallest subnetwork with width$\times0.25$ has the largest increase (0.8\%) from 73.6\% to 74.4\%.

\textit{Which is the best teacher for small submodels? }
We train our slimmable model with soft targets predicted by different models as teachers in knowledge distillation. For the student subnetwork $\mathcal{S}(\theta_{w_{n}})$, 'prev', 'largest', 'mean' indicates that the soft target is the predicted probability $p^{n+1}$ of the last larger subnetwork $\mathcal{S}(\theta_{w_{n+1}})$, $p^{N}$ of the largest subnetwork $\mathcal{S}(\theta_{w_{N}})$ \cite{yu2019universally} and the average of all the predictions $\frac{1}{N-n}\sum^{N}_{j=n+1}p^{j}$ by the subnetwork larger than the current model $\mathcal{S}(\theta_{w_{n+1}}),...,\mathcal{S}(\theta_{w_{N}})$, respectively. Different from the setting of 'mean', 'larger' represents using the average loss of all the larger submodels' distillation. The mIoU of our slimmable model under different teacher settings are reported in Table \ref{tab:teacher}. Note that all the models are trained with the sum of the three losses proposed. Our 'prev' setting, the stepwise downward distillation, outperform others by higher mIoU 74.4\% and 77.37\% on width $\times0.25$ and $\times0.5$. Using the average loss of all larger submodels results in better mIoUs on the larger submodels with width $\times0.75$ and $\times1.0$, but even lower mIoU than models trained without distillation on width $\times0.25$ and $\times0.5$. The results are consistent with the phenomenon that student network's performance degrades when the gap between student and teacher is too large \cite{mirzadeh2020improved}.

\begin{table}[htbp]
\caption{Ablation of knowledge distillation (KD) with different loss function by Slim-SFNet-ResNet50 on Cityscapes \textit{val}.}
% Using the distillation combined with boundary guided loss function achieve higher accuracy on all the subnetworks. KD indicates whether to use knowledge distillation.

\label{tab:distillation}
% \scalebox{1.0}{
\setlength{\tabcolsep}{1.0mm}{
\begin{tabular}{@{}c|ccc|ccc|cccc@{}}
\toprule
\multirow{2}{*}{KD} & \multicolumn{3}{c|}{GT} & \multicolumn{3}{c|}{Soft Target} & \multicolumn{4}{c}{mIoU  (\%)}\\ 
                     & $\mathcal{L}_{seg}$ & $\mathcal{L}_{b}$ & $\mathcal{L}_{g}$ & $\mathcal{L}_{seg}$ & $\mathcal{L}_{b}$ & $\mathcal{L}_{g}$ & $\times 0.25$ & $\times 0.5$ & $\times 0.75$ & $\times 1.0$ \\ \midrule
\multirow{4}{*}{w/o} & \Checkmark  &   &   &   &   &   &71.82   &75.97   &76.92  & 77.90 \\
                     & \Checkmark  & \Checkmark  &   &   &   &   &73.08   &76.34   &77.12   &78.14 \\
                     & \Checkmark  &   & \Checkmark  &   &   &   &72.49   &76.47   &77.82  &78.35\\
                     & \Checkmark  & \Checkmark  & \Checkmark  &   &   &   &73.63   &76.92   &77.77   &78.26\\ \midrule
\multirow{4}{*}{w}   & \Checkmark  &   &   & \Checkmark  &   &   &71.94   &75.86   &76.64   &77.55\\
                     & \Checkmark  & \Checkmark  &   & \Checkmark  & \Checkmark  &     & 73.12  & 76.04  & 77.21  & 78.21 \\
                    & \Checkmark  &  & \Checkmark   & \Checkmark  &   & \Checkmark  &72.94   &76.16  &77.41   &78.37 \\
                    & \Checkmark  & \Checkmark  & \Checkmark  & \Checkmark  & \Checkmark  & \Checkmark  &\textbf{74.40}   &\textbf{77.37}   &\textbf{77.87}   &\textbf{78.43}\\
                     \bottomrule
\end{tabular}}
% }
% \vspace{-2mm}
\end{table}

\begin{table}[htbp]
\vspace{-1mm}
\caption{Ablation of different knowledge distillation (KD) strategies with Slim-SFNet-ResNet50 on Cityscapes \textit{val}. \textbf{Bold} numbers and \textit{italic} numbers indicate the best and second best results.}
%  Using the stepwise downward distillation outperforms the others on smaller subnetworks. KD indicates whether knowledge distillation is used
% \vspace{-2mm}
% \scalebox{1.0}{
\label{tab:teacher}
\setlength{\tabcolsep}{0.5mm}{
\begin{tabular}{@{}c|c|c|cccc@{}}
\toprule
\multirow{2}{*}{KD} & \multirow{2}{*}{Teacher} & \multirow{2}{*}{Loss} & \multicolumn{4}{c}{mIoU (\%)} \\
                   &  & & $\times 0.25$ & $\times 0.5$ & $\times 0.75$ & $\times 1.0$ \\ \midrule
w/o                &-  &$\mathcal{L}_{CE/BCE}(p^{n},y)$  &73.63   &76.92   &77.77   &78.26  \\ \midrule
\multirow{4}{*}{w} &prev &$\mathcal{L}_{KD}(p^{n},p^{n+1})$ &\textbf{74.40}   &\textbf{77.37}   &\textit{77.87}   &\textit{78.43}  \\
                   &largest  &$\mathcal{L}_{KD}(p^{n},p^{N})$  &73.64  &76.72  &77.04 &78.38  \\
                   &mean  &$\mathcal{L}_{KD}(p^{n},\frac{1}{N-n}\sum^{N}_{j=n+1}p^{j})$ &73.24 & 76.25 & 77.53 & 77.85 \\
                   &larger  &$\frac{1}{N-n}\sum^{N}_{j=n+1}\mathcal{L}_{KD}(p^{n},p^{j})$ &73.25 & 75.87 & \textbf{78.02} & \textbf{78.61} \\ \bottomrule
\end{tabular}
}
% }
% \vspace{-2mm}
\end{table}

\textbf{Boundary Supervision. }
% We demonstrate the effectiveness of the boundary supervision and then present some qualitative segmentation results. 
As shown in Table \ref{tab:distillation}, with boundary detection loss $\mathcal{L}_{b}$, the mIoUs on all widths are improved, especially for the smallest submodels, with 1.2\% increase from 71.94\% to 73.12\%. For slimmable models trained without $\mathcal{L}_{b}$ but with the boundary guided segmentation loss $\mathcal{L}_{g}$, we use the binary boundary ground truth label as a mask to generate a masked probability map $p_{ms}$. The boundary guided segmentation loss with ground truth labels also helps on improving the mIoUs on all width. With all the losses together, we get the best performance on all the submodels.

\begin{figure}[htbp]
\centering
\includegraphics[scale=0.53]{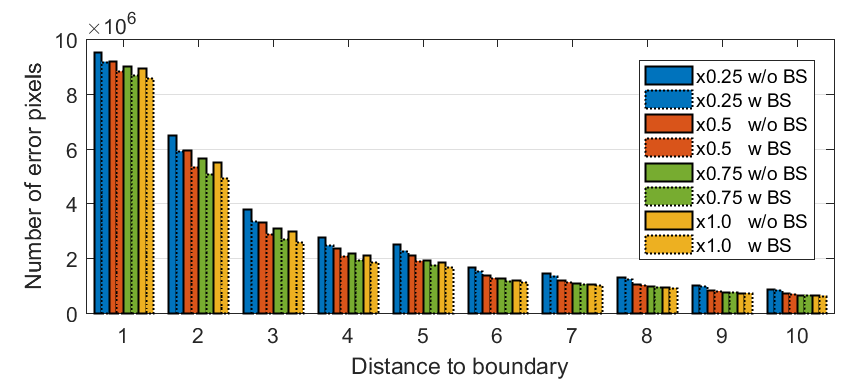}
\vspace{-2mm}
  \caption{Comparison of the distribution of error pixels between slimmable models trained with and without boundary supervision (BS) on Cityscapes \textit{val}. The model with boundary supervision has less error predictions on the boundary.}
  \label{fig:hist_error}
%   \vspace{-4mm}
\end{figure}

\begin{figure}[htbp]
\vspace{-2mm}
  \centering
  \includegraphics[scale=0.8]{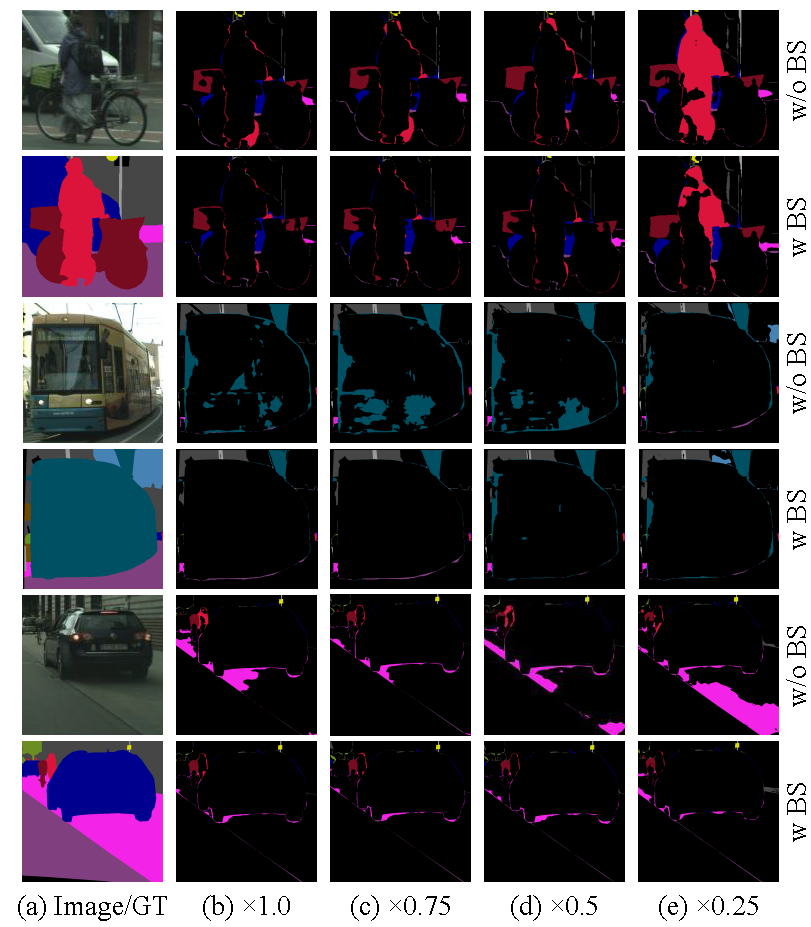}
%   \vspace{-2mm}
  \caption{Visual comparison of our boundary supervision (BS) on Cityscapes \textit{val}, in terms of errors in predictions, where correctly predicted pixels are shown as black background while wrongly predicted pixels are colored with their ground truth label color codes. Submodels with boundary supervision perform better on small objects and semantic borders.}
  \label{fig:edge_visual}
  \vspace{-3mm}
\end{figure}

To demonstrate the improvements on semantic borders, we illustrate the histogram of the error pixels in Figure \ref{fig:hist_error}. It shows the statistics of error pixels numbers and their Euclidean distances to the nearest boundaries on 500 Cityscapes \textit{val} images. Overall, the improved pixels are mainly distributed on the semantic borders. The improvement number of pixels within the range of 5 pixels along the borders accounts for about 50\% of the total. Some qualitative results on Cityscapes \textit{val} are shown in Figure \ref{fig:edge_visual}. 
With the boundary supervision, the predicted segmentation maps of each width model are more consistent, especially on the boundary regions. Segmentation results for some interior regions are also improved.

% \vspace{-3mm}
% \vspace{-3mm}

\subsection{Comparisons with Real-time Models}
We compare our method with other existing state-of-the-art real-time methods on Cityscapes and CamVid. 
% All the methods are evaluated by single-scale inference with the input image sizes listed for fair comparison. 
% Our speed is tested on one GTX 2080Ti GPU with full image resolution as input, and we report the speed of without TensorRT acceleration. 

\textbf{Results on Cityscapes. } We present the mIoU and inference speed of our slimmable SFNet-ResNet50 and SFNet-ResNet18 (both backbones are pretrained on ImageNet) and other real-time segmentation methods in Table \ref{tab:cityscapes_sota}. Our Slim-SFNet-ResNet50 achieves result (77.3\%) with FPS 23.8. With ResNet18 as backbone, our method achieves 74.3\% mIoU with 51.4 FPS.

% \vspace{-4mm}
\begin{table}[h]
\caption{Comparison with state-of-the-art on Cityscapes \textit{val}. \ddag indicates the model is not pretrained on ImageNet.}
\label{tab:cityscapes_sota}
% \scalebox{0.93}{
\setlength{\tabcolsep}{0.2mm}{
\begin{tabular}{l|c|c|c|c|c|c}
\toprule
\multicolumn{1}{c|}{Method} & Resolution & Backbone & mIoU & FLOPs & FPS & Param \\ \midrule
BiSeNetV1\cite{yu2018bisenet}          & 768$\times$1536     & Xception39       & 69.0      & 14.8        & 105.8 & 5.8      \\
BiSeNetV1\cite{yu2018bisenet}          & 768$\times$1536     & ResNet18          & 74.8     & 55.3        & 65.5  & 49       \\
CAS\ddag\cite{zhang2019customizable}                & 768$\times$1536    & Searched                & 71.6     & -           & 108   & -        \\
GAS\ddag\cite{lin2020graph}                & 767$\times$1537    & Searched                & 72.4    & -           & 163.9 & -        \\
DF1-Seg\cite{li2019partial}            & 1024$\times$2048    & DFNetv1               & 74.1     & -           & 106.4     & -        \\
DF2-Seg1\cite{li2019partial}            & 1024$\times$2048     & DFNetv2               & 75.9     & -           & 67.2     & -        \\
DF2-Seg2\cite{li2019partial}            & 1024$\times$2048    & DFNetv2               & 76.9     & -           & 56.3  & -        \\
% SFNet\cite{li2020semantic}              & $1024\times2048$    & DF2               & -        & 80.3        & 53    & 17.9     \\
SFNet\cite{li2020semantic}              & 1024$\times$2048     & ResNet18          & 78.7        & 247         & 18    & 12.9   \\
BiSeNetV2\ddag\cite{yu2021bisenet}           & 1024$\times$2048     & None                & 73.4     & 21.3       & -     & -        \\
BiSeNetV2-L\ddag\cite{yu2021bisenet}         & 512$\times$1024    & None                & 75.8     & 118.5      & 47.3  & 4.6     \\
FasterSeg\ddag\cite{chen2020fasterseg}         & 1024$\times$2048   & Searched                & 73.1    & 28.2        & 108.4 & 4.4      \\
STDC2-Seg75\cite{fan2021rethinking}        & 768$\times$1536     & STDC2             & 77.0     & 54.9        & 97\dag    & 16.1     \\
% MSFNet\cite{SiZL20}            & $512\times1024$      & ResNet18          & -        & 24.2        & 117   & -        \\
MSFNet\cite{SiZL20}             & 1024$\times$2048     & ResNet18          & 77.2       & 96.8        & 41    & -        \\
CABiNet\cite{yang2021real}            & 1024$\times$2048     & MBNetv3-s         & 76.6     & 12          & 76.5  & 2.64     \\
CABiNet\cite{yang2021real}             & 1024$\times$2048     & ResNet18          & 76.7      & 66.4       & 54.5  & 9.2     \\
DDRNet-Seg\cite{hong2021deep}      & 1024$\times$2048    & DDRNet-23         & 79.5        & 143.1       & 37.1  & 20.1     \\ \midrule
% DDRNet-Seg\cite{hong2021deep}      & $1024\times2048$    & DDRNet-39         & -           & 281.2       & 22    & 32.3     \\
% RegSeg\ddag\cite{gao2021rethink}             & 1024$\times$2048   & None              & 78.1       & 39.1        & 30    & 3.34     \\ \midrule
\multirow{4}{*}{\begin{tabular}[c]{@{}c@{}} Slim-SFNet \\ \footnotesize ${\times[0.25, 0.5, 0.75, 1.0]}$ \\ (Ours) \end{tabular}} & \multirow{4}{*}{1024$\times$2048}         & \multirow{4}{*}{ResNet50}      & 74.4     &  39.4   & 46.2                & 2.0   \\
  &          &          & 77.3  &153.9    & 23.8                & 7.8   \\
  &          &          & 77.8   & 343.4  & 13.2               & 17.6     \\
  &          &          & 78.4     &   607.9 & 9.0               & 31.2    \\ \midrule
  
\multirow{4}{*}{\begin{tabular}[c]{@{}c@{}} Slim-SFNet \\ \footnotesize ${\times[0.25, 0.5, 0.75, 1.0]}$ \\ (Ours) \end{tabular}}   & \multirow{4}{*}{1024$\times$2048}          & \multirow{4}{*}{ResNet18}    & 70.4    &15.7   & 74.9                & 0.8     \\
    &       &     & 74.3   &61.5   & 51.4              &  3.2  \\
    &       &     & 76.7   &137.4   & 30.8              &  7.2  \\
    &       &     & 77.9   &243.6  & 21.8               & 12.9   \\ 
\bottomrule

\end{tabular}}
% }
\vspace{-2mm}
\end{table}

\textbf{Results on CamVid. } Since the inference speed of different models is measured under different conditions, we list the corresponding GPU type. Table \ref{tab:camvid_sota} shows the comparison results on CamVid between our method and SoTA methods. Our network achieves competitive trade-off between performance and speed by $80.1\%$ ($72.5\%$ without ImageNet pretraining) mIoU with $55.7$ FPS, which outperforms the original independently trained SFNet.

\begin{table}[h]
\caption{Comparison with state-of-the-art on CamVid \textit{test} with image size 720$\times$960. IM and CS represent using extra data, ImageNet and Cityscapes, for pretraining, respectively. \dag indicates the FPS is measured with TensorRT acceleration.}
\label{tab:camvid_sota}

% \scalebox{0.93}{
\setlength{\tabcolsep}{0.8mm}{
\begin{tabular}{l|c|c|c|c|c}
\toprule
\multicolumn{1}{c|}{Method}   &Extra   &Backbone   &mIoU   &FPS  &GPU \\ \midrule
% ENet\cite{paszke2016enet}       & IM          & None         & 51.3  & 61.2   & TitanX    \\
% ICNet\cite{zhao2018icnet}       & IM          & PSPNet50   & 67.1   & 34.5  & TitanX     \\
BiSeNetV1\cite{yu2018bisenet}   & IM          & Xception39 & 65.6   & 175   & GTX1080Ti          \\
BiSeNetV1\cite{yu2018bisenet}   & IM          & ResNet18   & 68.7   & 116.3  & GTX1080Ti          \\
CAS\cite{zhang2019customizable}  & None       & Searched         & 71.2   & 169 & TitanXp           \\
GAS\cite{lin2020graph}           & None       & Searched         & 72.8   & 153.1  & TitanXp          \\
% SFNet\cite{li2020semantic}      & IM        & DF2        & 70.4  & 134  & GTX1080Ti      \\
SFNet\cite{li2020semantic}       & IM       & ResNet18   & 73.8  & 36  & GTX1080Ti       \\
MSFNet\cite{SiZL20}              & None       & None         & 75.4  & 91  & GTX2080Ti      \\
STDC1-Seg\cite{fan2021rethinking}    & IM     & STDC1      & 73.0   & 198\dag & GTX1080Ti  \\%TensorRT 5.0.1.5
STDC2-Seg\cite{fan2021rethinking}    & IM     & STDC2      & 73.9   & 152\dag  & GTX1080Ti    \\ 
BiSeNetV2\cite{yu2021bisenet}        & CS     & None         & 76.7    & 124.5  & GTX1080Ti    \\
BiSeNetV2-L\cite{yu2021bisenet}      & CS     & None         & 78.5    & 32.7  & GTX1080Ti       \\
% RegSeg\cite{gao2021rethink}          & CS     & None   & 80.9     & 70    & TeslaT4  \\ 
DDRNet-Seg\cite{hong2021deep}        & CS     & DDRNet-23         & 80.6       & 94   & GTX2080Ti          \\ \midrule

\multirow{4}{*}{\begin{tabular}[c]{@{}c@{}} Slim-SFNet \\ \footnotesize ${\times[0.25, 0.5, 0.75, 1.0]}$ \\ (Ours) \end{tabular}} & \multirow{4}{*}{CS}  & \multirow{4}{*}{ResNet50}   & 78.0     &57.1    & \multirow{4}{*}{GTX2080Ti}    \\
 &   &   & 80.6     &47.9    &      \\
 &   &   & 81.6     &31.7    &   \\
 &   &   & 81.7     &21.8    &      \\ \midrule

\multirow{4}{*}{\begin{tabular}[c]{@{}c@{}} Slim-SFNet \\ \footnotesize ${\times[0.25, 0.5, 0.75, 1.0]}$ \\ (Ours) \end{tabular}} & \multirow{4}{*}{IM}  & \multirow{4}{*}{ResNet18}   &71.0     &102.8      & \multirow{4}{*}{GTX2080Ti}     \\
 &   &   &73.6     &98        &    \\
 &   &   &74.8     &72.6      &      \\
 &   &   &75.2     &55.7      &      \\ \midrule
 
\multirow{4}{*}{\begin{tabular}[c]{@{}c@{}} Slim-SFNet \\ \footnotesize ${\times[0.25, 0.5, 0.75, 1.0]}$ \\ (Ours) \end{tabular}} & \multirow{4}{*}{CS}  & \multirow{4}{*}{ResNet18}   &75.0     &102.8      & \multirow{4}{*}{GTX2080Ti}     \\
 &   &   &77.9     &98        &    \\
 &   &   &79.5     &72.6      &      \\
 &   &   &80.1     &55.7      &      \\ 
\bottomrule
\end{tabular}}
% }
\vspace{-5mm}
\end{table}

\textbf{Discussion. }  
Our work tackles the design of efficient and adjustable segmentation methods. In contrast to the SoTA real-time semantic segmentation methods, the performance of our methods do not rely on well-crafted compact network architectures. The experimental results demonstrated that our method can be directly applied to the mainstream segmentation frameworks and turn the fixed-computation models into adjustable ones. In this work, we use globally consistent width multipliers, but the optimal width of can be different for each layer, so we believe that the accuracy-efficiency tradeoff still has room for improvement. Furthermore, combining with image content, input resolution and depth of the network, the dynamic inference can be further explored.

%===============================================================
\section{Conclusion}

In this paper, we propose a general slimmable semantic segmentation method, which enables adjustable accuracy-efficiency tradeoff through a width-swicthable segmentation network. We demonstrate the effectiveness of stepwise downward distillation on improving the performance of smaller subnetworks, and with less amount of features saved during training compared with other distillation strategies. Based on the observation of the difference between the predictions of each subnetwork, we introduce boundary supervision on low-level features of the network and propose a boundary guided loss to further improve the segmentation results of pixels along semantic borders. We demonstrate the effectiveness of the proposed method through extensive experiments with different mainstream semantic segmentation networks on the Cityscapes and CamVid. Our proposed method improves the accuracy of the smaller submodels without great accuracy drops on large submodels.

%===============================================================

\begin{acks}
We thank Dr. Javier Vazquez Corral for his valuable suggestions on the revision of this paper. This work was partially supported by National Science Foundation of China under Grant No.U19B2037 and No.61901384, Natural Science Basic Research Program of Shaanxi Province (Program No.2021JCW-03), Grant PID2021-128178OB-I00 funded by MCIN/AEI/10.13039/501100011033, ERDF “A way of making Europe” and the Ramón y Cajal grant RYC2019-027020-I. D.X. and P.W. thank the funding from China Scholarship Council (No.202006290209, No.201906290067).
\end{acks}
% We thank Dr. Javier Vazquez Corral for valuable comments on the paper. 

% \newpage

% \bibliographystyle{ACM-Reference-Format}
% % \balance
% \bibliography{main_paper}

% \newpage
% \appendix
% \label{appendix}
% \section{Algorithm}
% \subsection{Algorithm}

\end{document}